# SINGLE-IMAGE DRIVEN 3D VIEWPOINT TRAINING DATA AUGMENTATION FOR EFFECTIVE WINE LABEL RECOGNITION

[1]Yueh-Cheng Huang, [1]Hsin-Yi Chen, [1]Cheng-Jui Hung, [1]Jen-Hui Chuang, [2]Jenq-Neng Hwang

[1]Department of Computer Science, National Yang Ming Chiao Tung University, Taiwan
[2]Department of Electrical & Computer Engineering, University of Washington, United States

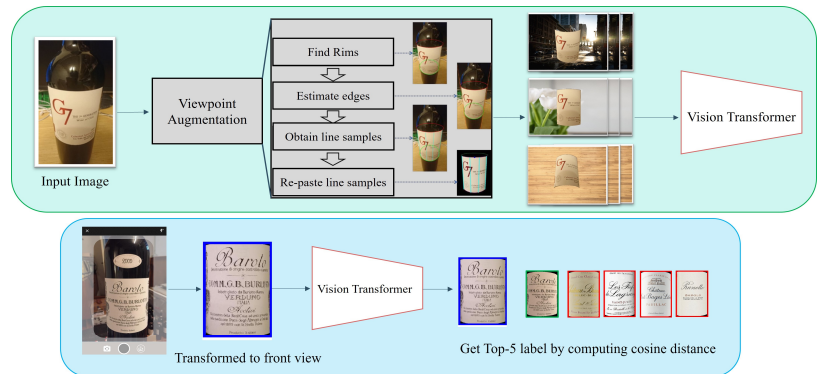

**Fig. 1.** Pipeline of wine label recognition with limited data

## ABSTRACT

Confronting the critical challenge of insufficient training data in the field of complex image recognition, this paper introduces a novel 3D viewpoint augmentation technique specifically tailored for wine label recognition. This method enhances deep learning model performance by generating visually realistic training samples from a single real-world wine label image, overcoming the challenges posed by the intricate combinations of text and logos. Classical Generative Adversarial Network (GAN) methods fall short in synthesizing such intricate content combination. Our proposed solution leverages time-tested computer vision and image processing strategies to expand our training dataset, thereby broadening the range of training samples for deep learning applications. This innovative approach to data augmentation circumvents the constraints of limited training resources. Using the augmented training images through batch-all triplet metric learning on a Vision Transformer (ViT) architecture, we can get the most discriminative embedding features for every wine label, enabling us to perform one-shot recognition of existing wine labels in the training classes or future newly collected wine labels unavailable in the training. Experimental results show a significant increase in recognition accuracy over conventional 2D data augmentation techniques.

## 1. INTRODUCTION

Automatic wine label recognition [1, 18] has become increasingly popular recently due to its practical usage. An intuitive way to accomplish the recognition task is by employing OCR (Optical Character Recognition) techniques to extract relevant text-only information. Similar to traditional approaches of offline handwriting recognition [14], many deep learning-based scene text detection methods [6, 7, 21-23] may be employed to identify text regions before subsequent recognition processes, as elaborated in [28]. However, this approach is susceptible to errors due to defaced text, which may negatively impact recognition accuracy [17]. Moreover, the variations in texts from different languages with mixed fonts and sizes, as well as tangled text and graphics [15], make this fine-grained recognition task even more challenging [25]. Therefore, using a computer vision model trained on images containing the entire wine label region [19, 20], not just the text, is considered in this paper.

On the other hand, the challenge of having insufficient training data remains a critical issue for such a model to perform satisfactorily. Regardless of the improvements in model design and training techniques, using insufficient and unrepresentative data for training can result in inadequate performance of generalization [8, 24]. Also, obtaining enough large and diverse training data that are representative of the target dataset remains a challenging task for many practical applications [11, 16, 27]. Moreover, when applying deep learning to real-world tasks, it is often encountered that the training data are very different from the test data.

Many works [10, 12] have shown that appropriate data augmentation techniques can help address this issue by generating additional training samples from the existing ones. Specifically, as described in [32], image data augmentation approaches can be roughly classified into two categories, which are: (i) based on basic image manipulations or (ii) based on deep learning. For (i), image augmentations can be carried out by geometric transformation (flipping, cropping, rotation, translation, shearing), noise injection, random erasing, color space transformation, and image mixing. Such manipulations try to preserve the main features of existing images from the training dataset, while adding potential variations for better generalization in the testing dataset. Similarly, deep-learning-based augmentations of (ii), such as adversarial training, style transfer, and generative adversarial networks (GANs), exploit CNN-based network(s) to achieve a variety of image styles, e.g., changing lighting directions and intensities, or generate realistic images through learned image features from the training dataset [2].

While existing data augmentation methods enhance testing accuracy in various learning-based tasks, they often fall short in more specialized applications [3]. Specifically, standard image manipulations fail to produce realistic images for certain tasks [30], while deep-learning-based augmentations necessitate additional, specific training images [5]. These limitations are especially pronounced in the task of wine label recognition, where conventional methods cannot adequately simulate the realistic perspective of labels on cylindrical wine bottles, while advanced methods require diverse images from multiple viewpoints for effective training[1].

To address the challenges in wine label recognition, this paper proposes an innovative 3D viewpoint augmentation pipeline, which generates a diverse and realistic training dataset from a single label image, to effectively train a vision transformer (ViT) for wine label recognition, as shown in Fig. 1. During the testing phase, as shown in the blue lower section of Fig. 1, the same viewpoint augmentation principle is applied to the given testing image to obtain a synthesized front-view wine label, which is then passed to the trained ViT to obtain the embedding features, which are then used to determine the Top-5 cosine similarities for label identification.

The ViT model with an MLP head is typically trained to classify input images into predefined categories. However, in the case of wine label recognition, the market is continuously introducing new varieties, often with subtle variations, such as different vintages [4]. A significant challenge with direct classification is the model's adaptability to these new labels without the need for retraining. Therefore, for wine label recognition, we employ metric learning based one-shot recognition, which involves comparing the similarity of feature vectors (embeddings) from the front-view test data with those of the original training data embeddings.

By combining 3D viewpoint training data augmentation for metric learning of embedding features, we have developed an efficient and precise wine label recognition system. This approach not only addresses the shortcomings of traditional methods in recognizing new varieties of wine labels but also demonstrates the tremendous potential of applying deep learning techniques in rapidly changing product categories. Contributions of the paper include:

(i) Our wine label recognition pipeline can be effectively trained with very limited training data, requiring as few as only one sample.

(ii) Our proposed 3D viewpoint data augmentation for metric learning can improve the Top-1 accuracy significantly, i.e., more than 14.6%, over the standard 2D data augmentation based deep learning model.

## 2. PROPOSED METHOD

### 2.1. 3D Viewpoint augmentation from a single image

In this section, the proposed 3D viewpoint data augmentation scheme for a single wine label image is presented. This scheme needs to estimate, to some extent, the corresponding pose of the cylindrical bottle, and generate perspective realistic texts and patterns of the wine label in the augmented image. As shown in the green upper section of Fig. 1, the 3D viewpoint augmentation consists of three critical steps:

**1. 2D Description of 3D Surface**: The process begins with converting the 3D cylindrical surface of the wine label into a 2D representation. This involves identifying both the upper and lower elliptical rims (latitudinal edges) and the two straight longitudinal edges.

**2. Line Sample Extraction**: Next, we use the vanishing points from the above longitudinal edges to extract 2D line samples along the label's longitudinal direction.

**3. Perspective Mapping**: The final step involves mapping these line samples onto an image of a cylindrical surface with a different pose. This mapping, which uses a view-invariant cross-ratio technique, ensures the correct perspective of the wine label on each line sample.

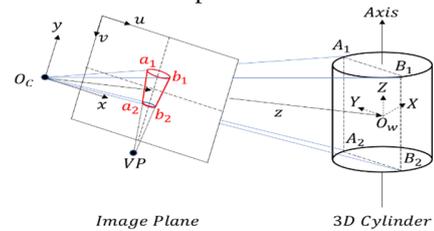

**Fig. 2.** Projective geometry of a cylinder

### 2.2. Projective geometry of a cylinder

According to projective geometry, images of the (circular) top and bottom plates of a (3D) cylinder, as illustrated in Fig. 2, will have elliptical shapes. As for the two edges of the cylinder in the image, i.e., $\overline{a_1 a_2}$ and $\overline{b_1 b_2}$, they correspond to the intersections of the image plane and planes $O_C A_1 A_2$ and $O_C B_1 B_2$, respectively, with both planes tangent to the 3D cylinder and passing through the camera center ($O_C$). Moreover, the intersection of $\overline{a_1 a_2}$ and $\overline{b_1 b_2}$ corresponds to the vanishing point (VP) of all 3D lines parallel to the axis of the cylinder.

---

[1] Although generative models can achieve good results in medical, painting, and other fields, they require large training data and cannot achieve data augmentation with a single training data.

### 2.3 Obtaining the image region of a rectangular wine label pasted on a bottle

In this section, a 2D geometric description of the image of a rectangular wine label pasted on a 3D cylindrical surface (of a wine bottle) is provided. Such description will be used in the next section to obtain 1D (longitudinal) line samples of the wine label region.

*2.3.1. Deriving upper and lower (elliptical) rims of the wine label region*

For the geometry shown in Fig. 2, elliptical expressions of the upper/lower rims of a roughly vertically oriented wine label region, e.g., for the image shown in Fig. 3 (a), can be obtained with the following procedure:

1. Convert the color image to a gray-level image (Fig. 3 (b))
2. Identify *edge pixels* with a large image gradient in the vertical direction (Fig. 3 (c))
3. Identify image blocks, i.e., *edge blocks*, with enough (relative to block dimension) edge pixels (Fig. 3 (d))
4. Label positive/negative (red/green) edge block according to the gradient direction of most edge pixels (Fig. 3 (e))
5. Establish the longest chains of positive, and negative, edge blocks (Fig. 3 (f))
6. Obtain (thinned) *rim pixels* by performing non-maximum suppression in the vertical direction (Fig. 3 (g))
7. Obtain *elliptical expressions* of the upper and lower rims via curve fitting (Fig. 3 (h))

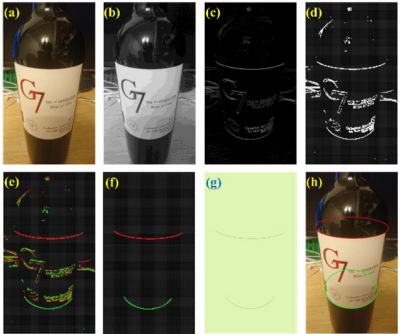

**Fig. 3.** Obtaining the elliptical expressions of the upper and lower rims of a wine label region (see text).

While the elliptical expressions in Step 7 can be obtained, for the rim pixels identified in Step 6, with an OpenCV function, default parameters need to be selected for some of the above processes, including minimum gradient (80, Step 2), block size (1/80 of image width, Step 3), minimum edge pixels (60% of block width, Step 3), and maximum gap in the chain (2 blocks, Step 5). Like a typical image processing procedure, different parameters may need to be determined, possibly manually, for some extreme imaging conditions. Nonetheless, such effort is worthwhile as unlimited synthetic, and perspective realistic, images can be obtained from such data augmentation, which will ultimately benefit the subsequent task of wine label recognition.

*2.3.2. Obtaining the left and right edges of the wine label*

As described in Sec. 2.2, left and right (longitudinal) edges of the wine label in an image correspond two common external tangents of the two ellipses obtained in Sec. 2.3.1. The procedure of finding these edges for two ellipses, e.g., $E_1$ and $E_2$ depicted in Fig. 4 (a), can be summarized as follows:

1. Identify search range for their intersection (VP) using bounding boxes of $E_1$ and $E_2$ ($Q_1$ to $Q_N$ in Fig. 4 (a)).
2. Obtain initial four tangents to $E_1$ and $E_2$ with $Q_1$ ($m_{11}$, $m_{12}$, $m_{21}$, and $m_{22}$ in Fig. 4 (b)).
3. Obtain the two common external tangents via binary search ($m_{11} = m_{21}$ and $m_{12} = m_{22}$ in Fig. 4 (c)).

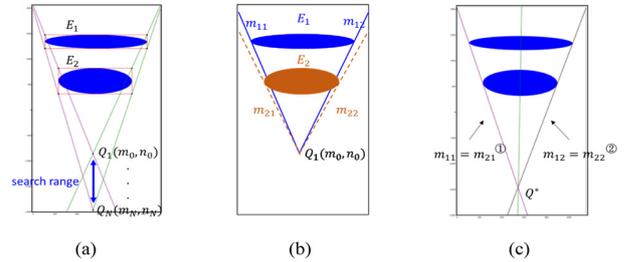

**Fig 4.** Obtaining left and right edges of a wine label region.

For Step 2, closed-form solutions of two tangents of an ellipse from an arbitrary point outside of the ellipse can be obtained; these are omitted here for brevity. As for Step 3, it is not hard to see that the slopes of all tangent lines will change monotonically with respect to the location of their intersection along a line and will not be the same except for a common external tangent; therefore, binary search can be employed to solve the problem efficiently.

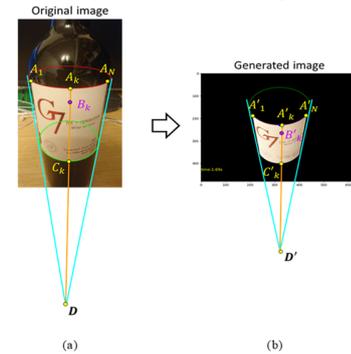

**Fig. 5.** (a) Obtaining (longitudinal) line samples of a wine label region, and (b) re-projecting them onto a virtual (invisible) wine bottle with an arbitrary pose (see text).

### 2.4 Obtaining 2D (longitudinal) line samples

To synthesize the image for a novel view of a wine label, point samples of the label need to be obtained from a real image captured in advance. To facilitate the geometrically natural synthesis process presented in the next subsection, these samples will first be obtained along the longitudinal direction of the wine label, i.e., is parallel to the axial direction of the wine bottle. As parallel lines in the 3D space will intersect at a vanishing point in an image, below is adopted to perform the foregoing sampling (see Fig. 5).

1. Identify the intersection of the two common external tangents as the vanishing point ($D$ in Fig. 5 (a)).
2. Identify the wider rim, and its leftmost/rightmost rim pixels ($A_1$ and $A_N$ in Fig. 5 (a)).
3. Obtain line samples by connecting rim pixels between $A_1$ and $A_N$ toward $D$, e.g., $\overline{A_k C_k}$ is identified as the $k$-th line sample with $C_k$ belonging to the smaller rim.

### 2.5 Synthesizing wine label images for perspective realistic data augmentation

Once foregoing line samples are obtained, they can be pasted onto the image of a cylindrical surface obtained from a novel view of the wine bottle, possibly via perspective projection of a graphic model. In this paper, such process is performed by pasting these line samples one at a time, with nonlinear mapping of pixel locations (based on the view-invariant cross-ratio between these locations) along each line according to the geometry of perspective projection, as shown in Fig 5 (b), so as to achieve perspective realistic appearance of the resultant synthetic image. In particular, the above re-projection process can be summarized as follows.

1. Identify the intersection of the two common external tangents as the vanishing point ($D'$ in Fig. 5 (b)).
2. Identify the wider rim, and its leftmost/rightmost rim pixels ($A'_1$ and $A'_N$ in Fig. 5 (b)).
3. Re-project image pixels of each line sample, e.g., $\overline{A_k C_k}$ in Fig. 5 (a), to the corresponding line segment connecting the two new rims, e.g., the $k$'-th line segment $\overline{A'_k C'_k}$ in Fig. 5 (b), using the cross-ratios.

While Steps 1 and 2 are like their counterparts in Sec. 2.4, locations of $A'_1$ and $A'_N$ in Fig. 5 (b), and thus the number of line samples need to be synthesized in Step3, need to be determined. As we have just a single image of wine label, i.e., the image shown in Fig. 5 (a), without having other camera/environmental information, these two locations are approximately estimated with respect to the width of the larger rim.[2] As for the re-projection performed in Step 3, since the location of three points are already determined along $\overline{A'_k C'_k}$, any image pixel of $\overline{A'_k C'_k}$ can be determined by solving the following equation of view-invariant cross-ratio, with $B'_k$ being the only unknown.

$$\frac{A'_k C'_k \cdot B'_k D'}{A'_k D' \cdot B'_k C'_k} = \frac{A_k C_k \cdot \boldsymbol{B_k} D}{A_k D \cdot \boldsymbol{B_k} C_k} \quad (1)$$

As the geometry of perspective projection is approximately satisfied in the foregoing process of re-projection, numerous visually realistic images of wine label can be generated from a real wine label image. Fig. 6 shows some synthetic images thus obtained from a single wine label image; wherein only one-dimensional rotation/translation of the virtual wine bottle is considered in each image so that the variation of its pose can be observed more easily. Note that the foregoing results are based on a virtual camera system which is established to mimic the imaging process of a typical cell phone camera. In particular, a virtual wine bottle with diameter equal to 76mm is placed about 150mm in front of the camera which has a focal length of about 6.8mm.

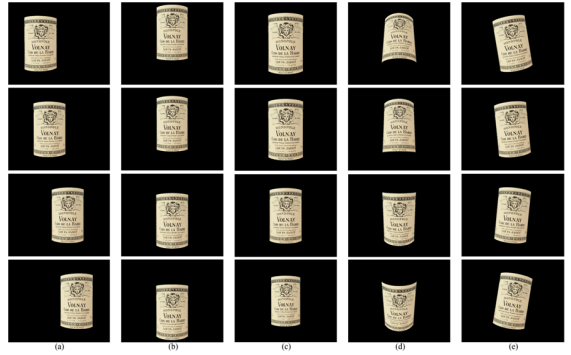

**Fig. 6.** Visually realistic (640×480) images synthesized upon a virtual (invisible) wine bottle which is translated between (a) x=040mm and x=40mm, (b) y=020 mm and y=20mm, and (c) z=230 mm and z=270 mm and rotated between (d) 030° and 30° w.r.t. the x-axis, and (e) 010° and 10° w.r.t. the z-axis.

### 2.6 Embedding Features from Metric Learning of a ViT

Recently, ViT [13] has achieved superior results in computer vision compared to traditional CNN-based approaches. Its ability to exploit global contextual information, coupled with its strong representation learning capabilities, makes it particularly suitable for wine label classification tasks.

*2.6.1. Training and Testing procedure*

During training stage of ViT, we employ 3D viewpoint augmentation to augment our training data, obtaining 2D images of wine labels observed from different perspectives. Second, we train the model using the augmented images through metric learning to obtain the discriminative embeddings of wine labels. The objective of the model is to minimize the cosine distance between embeddings of wine labels of the same class, while simultaneously increasing such distance between embeddings of different classes. For the testing stage, our model primarily relies on the embedding feature representation of a single 3D viewpoint generated front-view image, even though we have expanded the dataset through data augmentation. Therefore, during similarity calculations, we do not compare with all the embeddings or the average embedding of all augmented training data. Instead, we focus on comparing with the embedding of the single front-view 3D viewpoint generated image. Moreover, considering the importance of the original data's quality on model performance, we use method mentioned above to ensure testing data are compared with the embedding of a 3D viewpoint generated front-view image. This method is particularly crucial for handling wine labels with subtle variations, as it allows the model to accurately identify new or slightly altered labels based on a reliable and consistent reference point.

---

[2] Although further investigation is still needed for such issue, images synthesized with these simple estimations seem to work satisfactorily, as will be demonstrated in the experimental results.

*2.6.2. ViT Dino and loss function*

In our study, we have adopted the ViT architecture [13] which has been further advanced in the context of self-supervised learning within the DINO framework by Caron et al. [9]. Our approach is in line with the implementation utilized in DeiT [29], known for its effectiveness across a range of image processing tasks. For loss function, we use batch all triplet loss strategy proposed by [14], which is a variation of the conventional triplet loss [26].

With triplet loss, given an anchor sample $x_a$, the projection distance $D$ of a positive sample $x_p$ belonging to the same class $x_a$ should be closer to the anchor's projection than that of a negative sample belonging to a different class $x_n$, by at least a margin $m$. On the other hand, the batch all triplet loss aims to enhance the efficiency and effectiveness of training deep metric learning models. The batch all triplet loss, denoted as $\mathcal{L}_{BA}$, involves forming batches by randomly selecting $P$ classes (wine identities) and randomly sampling $K$ images from each class (wine). Then, it computes the triplet loss for all possible combinations of triplets, given by:

$$\mathcal{L}_{BA}(\theta;X) = \overbrace{\sum_{i=1}^{P}\sum_{a=1}^{K}}^{all\ anchors} \overbrace{\sum_{\substack{p=1\\p\neq a}}^{K}}^{all\ pos.} \overbrace{\sum_{\substack{j=1\\j\neq i}}^{P}\sum_{n=1}^{K}}^{all\ negatives} \left[m + d_{j,a,n}^{i,a,p}\right]_+$$

where

$$d_{j,a,n}^{i,a,p} = D\left(f(x_a^i), f(x_p^i)\right) - D\left(f(x_a^i), f(x_n^j)\right).$$

## 3. EXPERIIMENTS

### 3.1 Dataset

The dataset consists of 885 unique wine labels, with each label having only one corresponding label-on-bottle image in the training set (e.g., Fig. 7(a)). These images are collected and cropped from the website of wine companies. Consequently, the available data is insufficient to train a conventional model well. The majority of these training images consist of wine salon photographs, which are unbiased and frontal photos of wine bottles, taken without much inclination, as illustrated in Fig. 7 (a). As for the testing set, a total of 3,694 images of wine label are captured with an ordinary cell phone, for the representatives from the 885 classes. Each wine label is represented by 3 to 5 images, which exhibit various characteristics such as complex backgrounds, different lighting conditions, potential shadows, and wine bottle poses, as shown in Fig. 7 (b).

For the training data, there are 885 wine label images (one for each class), and a total of 320 images are generated for each image with our 3D viewpoint augmentation method by randomly rotating the virtual wine bottle between −10° and 10° w.r.t the z-axis, between −15° and 15° w.r.t the x-axis, and randomly translating the bottle by −40 to 40mm, −20 to 20mm, and 230 to 270mm in the x, y, and z directions, respectively. For the 2D augmentation method, a rotation between −15° and 15°, a relative translation between −5% to 5%, and a perspective transformation are again used to generate augmented image data. In addition, both methods incorporate randomly changing RGB color channels, image contrast/brightness, and color saturation. Finally, all augmented images are resized to a resolution of 224×224 pixels.

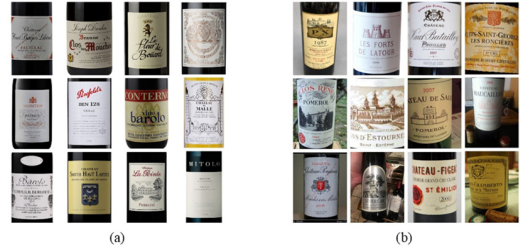

(a) (b)

**Fig. 7.** Datasets: Some image instances of wine label in (a) training dataset and (b) testing dataset

### 3.2 Improvement achieved through viewpoint augmentation

Firstly, we evaluated the performance of various deep learning models, including various ResNet and ViT architectures, on the wine dataset using conventional 2D data augmentation as well as our 3D viewpoint augmentation. Table. 1 provides a comprehensive evaluation of the wine label recognition accuracy.

It is readily observable that the visually more realistic 3D viewpoint augmentation outperforms the 2D data augmentation by a large margin, i.e., up to 14.91% improvement for the ViT-S/16 model's Top-1 results, indicating that more relevant image embeddings can be generated with our 3D augmentation scheme.

**Table 1.** Performance comparison of 2D traditional augmentation method and our 3D viewpoint augmentation method across different models.

| Condition | 2D Augmentation | | 3D Augmentation | |
|---|---|---|---|---|
| | Top-1 Acc. | Top-5 Acc. | Top-1 Acc. | Top-5 Acc. |
| VIT-S/16 | 76.39% | 83.26% | **91.15%** | **92.31%** |
| VIT-S/8 | 72.70% | 81.13% | 89.79% | 91.08% |
| VIT-B/16 | 75.29% | 83.26% | 91.02% | 92.11% |
| ResNet-50 | 72.46% | 78.19% | 76.44% | 80.16% |
| ResNet-101 | 71.92% | 78.80% | 72.79% | 80.10% |

When comparing ViT and ResNet with the inclusion of the proposed data augmentation and metric learning, we also observe a significant performance improvement achieved by our viewpoint augmentation on ViT, surpassing that of ResNet by a considerable margin. Specifically, taking ViT-S/16 and ResNet-50 as examples, the former actually outperforms the latter model by 14.76%. This substantial performance gap can be attributed to the fact that the embeddings generated by ViT possess higher discriminative power compared to those produced by ResNet, as depicted by the heatmap in Fig. 8, where one can see that the embeddings generated by ResNet for different input images tend to concentrate in the middle region, whereas the embeddings produced by ViT show distinct emphasis on various textures for different images. Consequently, during the process of image classification through metric learning, the embeddings generated by ResNet are more similar in terms of weights between different classes, leading to higher susceptibility to misclassification.

Finally, we can find that ViT-S/16 model achieved the highest Top-1 accuracy performance of 91.15%, which indicates that it is the most effective model for wine label recognition tasks. Based on such results, we plan to conduct further experiments and investigations specifically focused on the ViT-S/16 model.

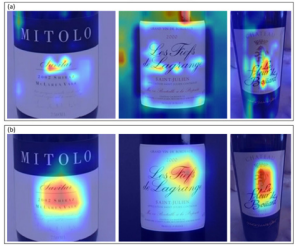

**Fig. 8.** Heatmaps obtained from ViT-S/16 (top) and ResNet-50 (bottom) showing embeddings for various input images.

### 3.3 Enhancing wine label recognition with background replacement

3D viewpoint augmentation involves augmenting the wine bottle data by introducing different poses of the bottles. However, it only generates foreground wine labels, while the background remains unspecified (black). Having a purely black background not only fails to fully exploit the data synthesis characteristics but also increases the risk of model overfitting. Therefore, we replace the black background with randomly sourced background images from the internet, as shown in Fig. 8. It is shown in Table 2, after the such replacement, the accuracy of recognition may increase by 3.27%. The reason behind such an improvement is that complex backgrounds introduce additional variations and noises into the data. This challenges the model to discern relevant foreground features and learn to focus on the important aspects of the input. By learning to ignore or adapt to complex backgrounds, the model becomes more robust in distinguishing signal from noises.

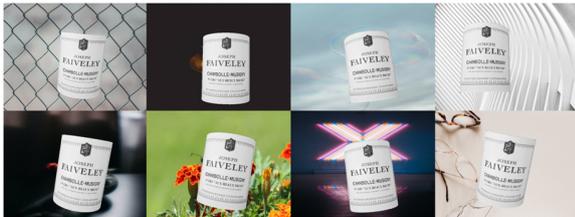

**Fig. 8.** Image synthesis examples with black background regions replaced by random background images.

### 3.3.1 Comparative analysis of perspective transform

2D Perspective transform is a commonly used data augmentation technique that allows images to be transformed from one perspective to another. Similar to our approach, it can alter the position, angle, and size of objects in an image to simulate the viewpoint of an observer to some extent. Nonetheless, due to the special curved surface of wine labels cropped from wine bottles, our 3D viewpoint augmentation can create more realistic augmentation of wine label images than those created by 2D perspective transform.

**Table 2.** Accuracy of ViT-S/16 in wine label recognition using 3D viewpoint augmentation, and for different backgrounds.

| ViT-S/16 + 3D Aug. | Condition | Top-1 Acc. | Top-5 Acc. |
|---|---|---|---|
| | w/o Background Replacement | 87.88% | 89.90% |
| | w/ Background Replacement | **91.15%** | **92.31%** |

To show the performance difference of 2D perspective transform compared with our method, we conduct experiments on the ViT-S/16 model for different degrees of perspective transforms. As shown in Table 3, where all our 3D viewpoint augmentation results outperform the 2D augmentation with different settings of perspective transform. Interestingly, while adding more 2D perspective transformations will indeed improve the accuracy for 2D augmentation methods in the ViT-S/16 model, it may actually have negative impacts on the performance of our 3D augmentation results. For example, our best results for the Top-1 accuracy is achieved by skipping the 2D perspective transformation completely.

**Table 3.** Accuracy of wine label recognition using ViT-S/16 and different perspective transformation schemes.

| Condition | 2D Augmentation | | 3D Augmentation | |
|---|---|---|---|---|
| | Top-1 | Top-5 | Top-1 | Top-5 |
| ViT-S/16 + Little Perspective Aug. | 76.79% | 84.44% | 91.15% | **92.31%** |
| ViT-S/16 + Big Perspective Aug. | 83.52% | 86.66% | 89.68% | 90.94% |
| ViT-S/16 + No Perspective Aug. | - | - | **91.57%** | 91.97% |

## 4. CONCLUSION

Data augmentation is a way to extend training data so that deep learning models can achieve good results in situations where such data are limited, of poor quality, or even absent. In this paper, such a problem is investigated for the task of automatic wine label recognition, and a novel data (3D viewpoint) augmentation technique is proposed to generate visually realistic training images, for essentially unlimited number of wine bottle poses, from a single wine label image captured in the real world. Experimental results show that the proposed augmentation technique can significantly improve the performance of the task of wine label recognition models, by 14.6% over the traditional 2D image data augmentation, when the training data is extremely limited, e.g., having only one image for each wine class.